\documentclass[12pt]{iopart}
\makeatletter
\@namedef{ver@amsmath.sty}{}
\makeatother
\usepackage{chemformula}
\usepackage{url}
\usepackage{amssymb,amsfonts}
\usepackage{amsmath}
\usepackage{graphicx}
\usepackage{bm}
\usepackage{multirow}
\usepackage{cuted}
\usepackage{subcaption}
\usepackage{flushend}
\usepackage{comment}
\usepackage[flushleft]{threeparttable} % http://ctan.org/pkg/threeparttable
\usepackage{booktabs}

\begin{document}

\title[]{CT respiratory motion synthesis using joint supervised and adversarial learning}

\author{Y-H Cao$^1$, V Bourbonne$^{1,2}$, F Lucia$^{1,2}$, U Schick$^{1,2}$, J Bert$^{1,2}$, V Jaouen$^{1,*}$ and D Visvikis$^{1,*}$}

\address{$^1$ LaTIM, UMR Inserm 1101, Université de Bretagne Occidentale, IMT Atlantique, Brest, France}
\address{$^2$ CHRU Brest University Hospital, Brest, France}
\address{$^*$ D. Visvikis and V. Jaouen share senior authorship}
\ead{vincent.jaouen@imt-atlantique.fr}
%\vspace{10pt}

\begin{abstract}
% As part of your revisions, please structure your abstract under the following headings: Objective, Approach, Main results, Significance.

Objective: Four-dimensional computed tomography (4DCT) imaging consists in reconstructing a CT acquisition into multiple phases to track internal organ and tumor motion. It is commonly used in radiotherapy treatment planning to establish planning target volumes. However, 4DCT increases protocol complexity, may not align with patient breathing during treatment, and lead to higher radiation delivery. 
Approach: In this study, we propose a deep synthesis method to generate pseudo respiratory CT phases from static images for motion-aware treatment planning. The model produces patient-specific deformation vector fields (DVFs) by conditioning synthesis on external patient surface-based estimation, mimicking respiratory monitoring devices. A key methodological contribution is to encourage DVF realism through supervised DVF training while using an adversarial term jointly not only on the warped image but also on the magnitude of the DVF itself. This way, we avoid excessive smoothness typically obtained through deep unsupervised learning, and encourage correlations with the respiratory amplitude.
Main results: Performance is evaluated using real 4DCT acquisitions with smaller tumor volumes than previously reported. Results demonstrate for the first time that the generated pseudo-respiratory CT phases can capture organ and tumor motion with similar accuracy to repeated 4DCT scans of the same patient. Mean inter-scans tumor center-of-mass distances and Dice similarity coefficients were $1.97$mm and $0.63$, respectively, for real 4DCT phases and $2.35$mm and $0.71$ for synthetic phases, and compares favorably to a state-of-the-art technique (RMSim).   
Significance: This study presents a deep image synthesis method that addresses the limitations of conventional 4DCT by generating pseudo-respiratory CT phases from static images. Although further studies are needed to assess the dosimetric impact of the proposed method, this approach has the potential to reduce radiation exposure in radiotherapy treatment planning while maintaining accurate motion representation. Our training and testing code can be found at \url{https://github.com/cyiheng/Dynagan}.

%Four-dimensional computed tomography (4DCT) consists in reconstructing an acquisition into multiple phases to track internal organ and tumor motion. It is commonly used in radiotherapy treatment planning but leads to higher doses of radiation, up to six times a conventional 3D CT scan. In this study, we propose a deep image synthesis method to generate pseudo respiratory CT phases from static images for motion-aware, 4DCT-free treatment planning. The model produces patient-specific deformation vector fields (DVFs) by conditioning synthesis on external patient surface-based estimation, mimicking respiratory monitoring devices. A key methodological contribution is to encourage DVF realism through supervised DVF training while using an adversarial term jointly not only on the warped image but also on the magnitude of the DVF itself. This way, we avoid excessive smoothness typically obtained through deep unsupervised learning, and encourage correlations between the respiratory amplitude and the generated image.
%We validate our model using real 4DCT acquisitions by evaluating global lung motion and image fidelity but also tumor displacements and motion of neighboring organs at risk. Importantly, we show that differences with ground truth are in the range of those achieved using real repeat 4DCT acquisitions. Our training and testing code can be found at \url{https://github.com/cyiheng/Dynagan}

\end{abstract}

\noindent{\it Keywords}: Image synthesis, GAN, dynamic imaging, respiratory motion, radiotherapy
%\maketitle

\section{Introduction}

% Intro sur l'utilité des CT, 4DCT & des Target Volumes -> Mais soucis avec la dose de radiation -> Besoin d'alternative
Four-dimensional computed tomography (4DCT) consists in reconstructing a 3D CT acquisition into multiple time frames to follow internal motion \cite{cox_image-guided_2007}. It is used routinely for radiotherapy (RT) treatment planning of lung cancer to ensure dose coverage to the tumor while reducing margins for better sparing of surrounding healthy tissues. A 4DCT acquisition is generally achieved using an external motion synchronization device (such as a respiratory belt) to average the CT signal of several phases -or gates- of the respiratory cycle, a process known as respiratory gating \cite{liu_evaluation_2018}. According to RT terminology, the \textit{Internal Target Volume} (ITV) including all potential locations of the \textit{Gross Tumor Volume} (GTV) during breathing is obtained from the 4DCT scan. After the addition of set-up margins, a \textit{Planning Target Volume} (PTV) is used for the treatment planning \cite{sgouros_icru_2021}. 

The benefits of using 4DCT acquisition over conventional CT for ITV delineation have been reported in many studies \cite{underberg_four-dimensional_2004, guckenberger_is_2007,louie_inter-observer_2010}. However, it also suffers from several drawbacks. Aside from practical issues such as cost and additional protocol complexity, the longer acquisition times result in much higher radiation exposure, up to six times a conventional 3D CT \cite{van_sornsen_de_koste_renal_2006}. 

\begin{figure}[t]
    \centering
    \includegraphics[width=0.85\columnwidth]{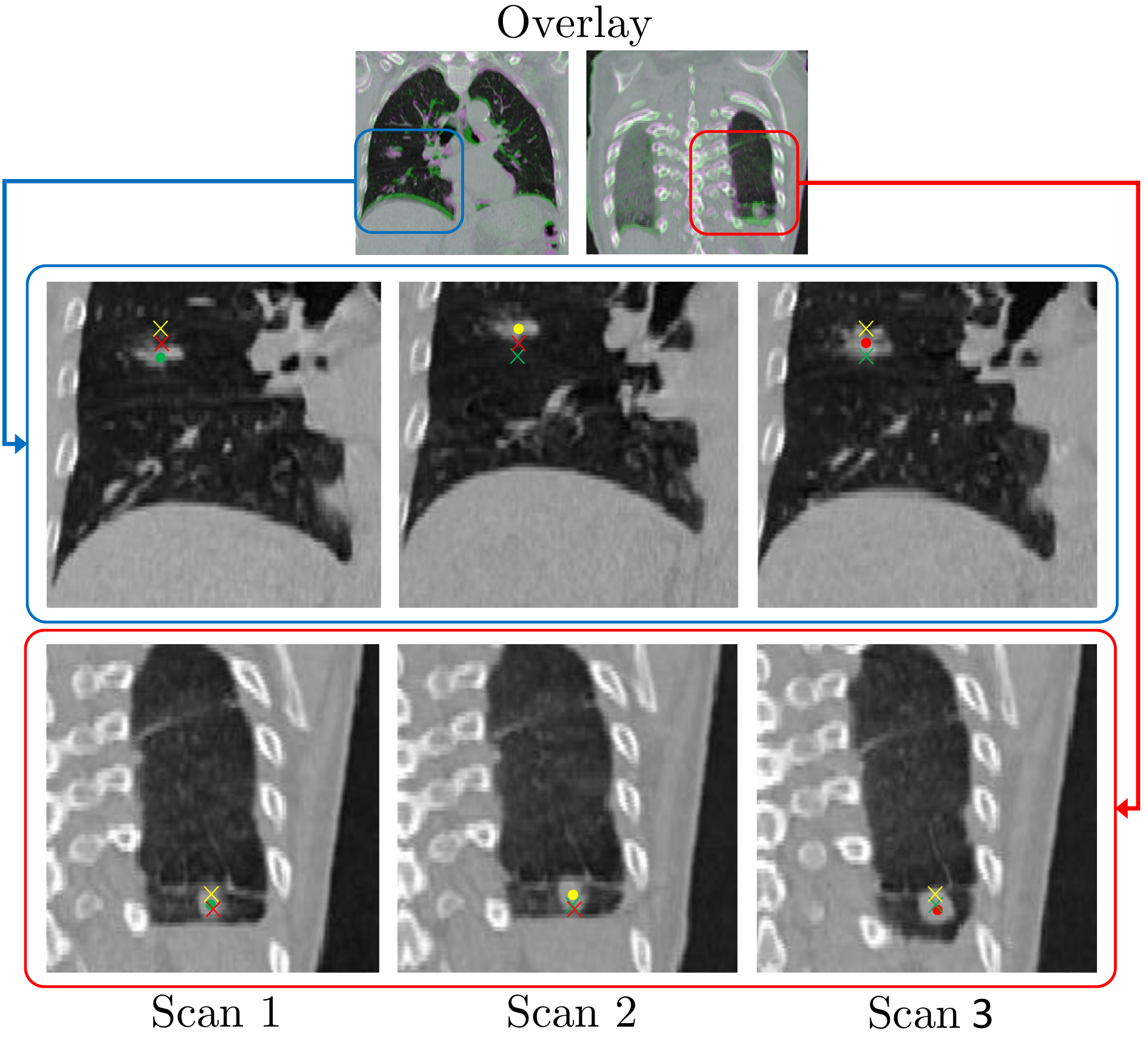}
    % \captionsetup{justification=centering}
    \caption{ Variability of 4DCT tumor tracking. Here, two patients underwent multiple repeat 4DCT scans due to excess variability between corresponding phases. Green, yellow, and red dots indicate the center of the tumor in scans 1, 2, and 3 respectively. Colored crosses show the corresponding tumor position in the other scans. Images from CHRU Brest.}
    \label{intravar}
\end{figure} 

Another important limitation of 4DCT is that the actual respiratory pattern during the RT session may differ due to the inherent irregular nature of respiration. It is thus common to perform repeat 4DCT scans to assess this variability and capture the full extent of tumor motion (with usually greater confidence given to the second 4DCT scan due to a better familiarity of the patient with the protocol \cite{de_oliveira_duarte_use_2019,tremolieres_lung_2022}). Repeat scans can even be performed more than once if the tumor position between identical gates differs exceedingly (Fig. \ref{intravar}), which is relatively common in the lower lobes \cite{sarudis_systematic_2017, vergalasova_modern_2020}. Nevertheless, despite these limitations, 4DCT is a key component of modern lung RT due to improved accuracy of target delineation and better healthy tissue sparing \cite{thariat_past_2013}. There is therefore a clear need to achieve similar dosimetric objectives with more optimized protocols in terms of imaging-induced dose and clinical workflow complexity. 

In this work, we propose a method to generate patient-specific pseudo respiratory motion synchronized chest CT images from static 3D CT scans and external surface-based estimation using deep image synthesis. Contrary to DIR, the objective is to synthesize motion from a single image (i.e. without fixed/moving image pairs). To this end, we propose several key contributions: 

\begin{itemize}
    \item  At the architectural level, we generate patient-specific deformation vector fields (DVFs) during breathing by conditioning a 3D U-Net encoder-decoder to the respiratory amplitude, that can be obtained with any external respiratory monitoring system. This is done through adaptive instance normalization of the bottleneck of the encoder, a technique usually used for style transfer \cite{huang_arbitrary_2017}. At test-time, the proposed model synthesizes any desired respiratory phase from a static 3D CT image. This approach allows for the creation of dynamic, time-varying images that can model motion of internal lesions and organs during breathing.  
    
    \item Contrary to existing approaches derived from unsupervised deformable image registration (DIR) that use reconstruction losses and explicit DVF smoothness constraints during model training, we propose to supervise our model using highly accurate lung DVFs and to jointly penalize the warped image and the DVF magnitude through an additional adversarial objective. Doing so, we show substantial performance improvements over more conventional losses, achieving more faithful synthetic motion patterns. 

    \item We validate our approach extensively on real 4DCT images, not only by measuring the global motion of the lungs using volumetric overlaps like previous studies, but also by considering tumors and neighboring organs at risk. In particular, we show that the agreement between synthesized and real 4DCT phases is similar to that obtained between two real repeat 4DCT acquisitions of the same patient. Using a private dataset in addition to a public one reported in previous studies, we also stress the importance of considering smaller tumor populations for model validation. 
    
    \item For reproducibility and ease of use, we provide our training code and a graphical application with pre-trained model weights that can produce dynamic time series from any static lung 3D CT image. Our code is available at \url{https://github.com/cyiheng/Dynagan}. 
  
\end{itemize}

\section{Related works}
\begin{figure}[htbp]
    \centering
    \begin{subfigure}[b]{0.35\columnwidth}
        \centering
        \includegraphics[width=\textwidth]{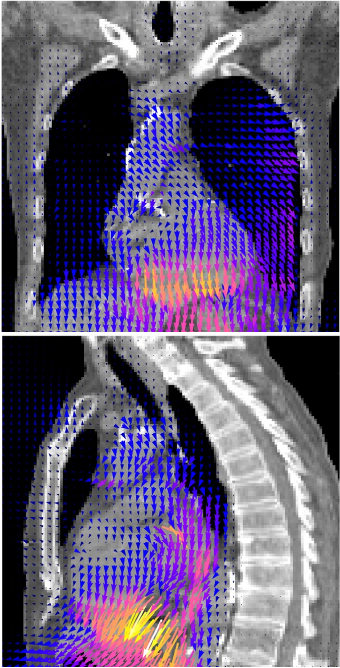}
        \caption{Conventional DIR}
        \label{Trad_DVF}
    \end{subfigure}
    \begin{subfigure}[b]{0.35\columnwidth}
        \centering
        \includegraphics[width=\textwidth]{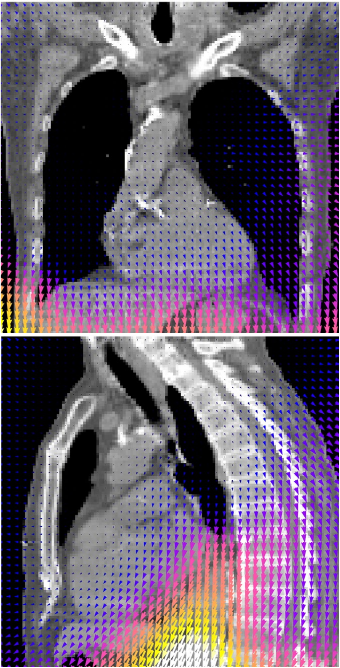}
        \caption{DL-based DIR}
        \label{DL_DVF}
    \end{subfigure}
    \caption{Comparison of typical appearances of deformation vector fields obtained from state-of-the art (a) conventional lung DIR \cite{vishnevskiy_isotropic_2017} and (b) deep learning-based lung DIR \cite{hansen_graphregnet_2021}.}
    \label{dir} 
\end{figure} 

\begin{figure}[t]
    \centering
    \includegraphics[width=\textwidth]{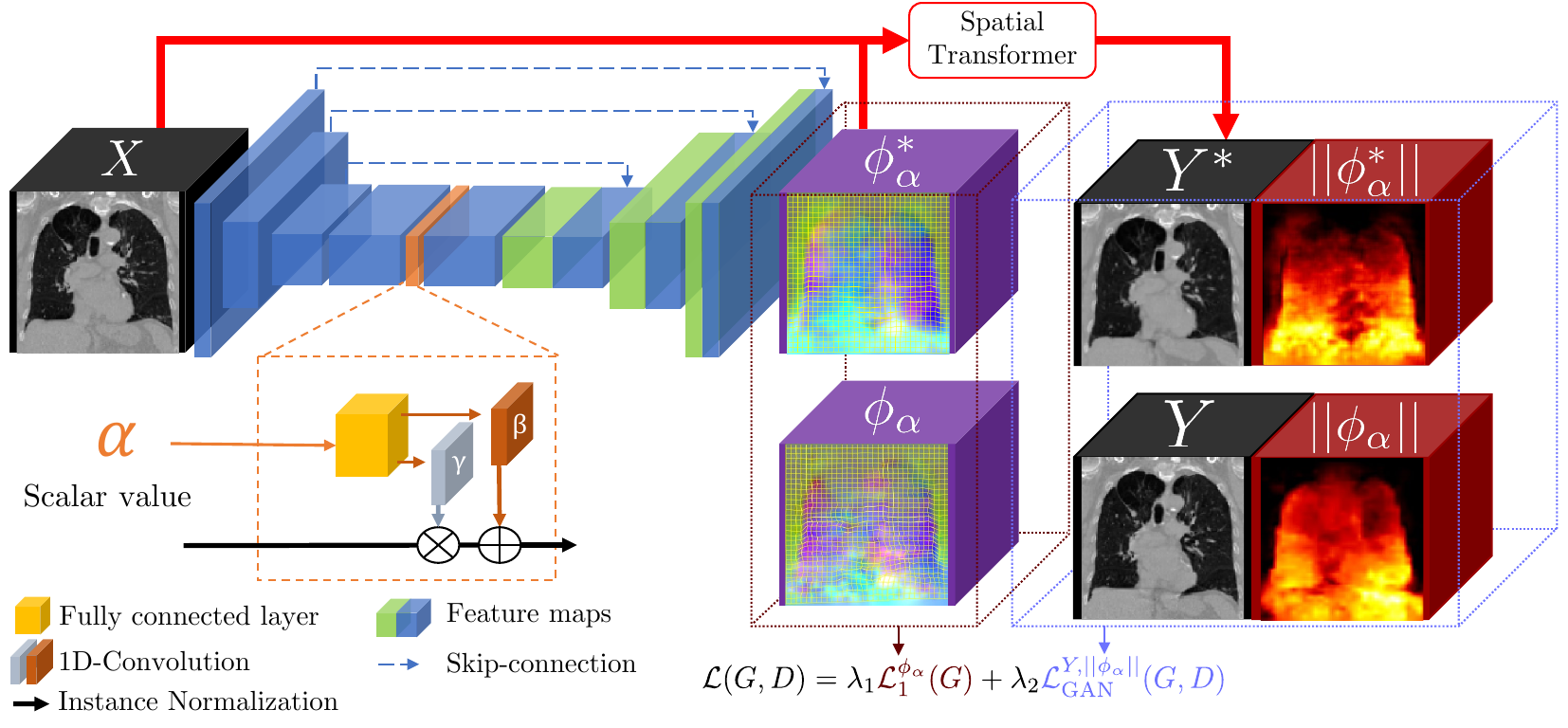}
    % \captionsetup{justification=centering}
    \caption{Proposed convolutional DVF generating architecture with scalar conditioning. The respiratory amplitude is injected through an AdaIN mechanism at the bottleneck of the encoder. Given an input image $X$ and a respiratory amplitude $\alpha$, the model learns a DVF $\phi^*_\alpha$ from supervised labels $\phi_\alpha$. Then the input image is warped with a spatial transformer to obtain the synthetic phase $Y^* = X\circ \phi^*_\alpha$. An adversarial loss based on the concatenation of the warped image and the magnitude of the DVF $||\phi_\alpha||$ is used to improve the realism of the DVF without explicitly constraining its smoothness.
    }
    \label{flow} 
\end{figure}

Correlations between patient surface respiratory signal and internal organ motion were reported in previous studies \cite{schweikard_respiration_2004, fayad_technical_2011, fayad_patient_2012}. These correlations have been used to build motion models for the estimation of internal organ motion from surrogate data \cite{mcclelland_respiratory_2013}. Different approaches to model motion have been described in the literature, for example, using diaphragm displacement \cite{Cervino_2009} for tumor motion or skin surface data for the estimation of internal respiratory motion \cite{McClelland_2011}. % Correlations between patient surface respiratory signal and internal organ motion were reported in previous studies \cite{ schweikard_respiration_2004, fayad_technical_2011, fayad_patient_2012} and exploited in population-based analysis to model respiratory motion, mainly using principal component analysis.
Fayad \emph{et al.} \cite{fayad_generation_2013} generated synthetic 4DCT phases by interpolating DVFs between end-of-inhale (EOI) and end-of-exhale (EOE) phases obtained by DIR. Based on similar concepts, Huang \emph{et al.} \cite{huang_deep_2022} predicted the principal components of respiratory DVFs with neural networks using time-of-flight surface information and a conventional lung DIR method \cite{vishnevskiy_isotropic_2017}. Both these works however require a 4DCT acquisition during deployment. Other works focus on 3D CT to 4DCT image generation using unsupervised DL-based DIR between training phases in order to apply the transformation to a static 3D image at test-time. Chang \emph{et al.} \cite{chang_generative_2021} proposed a method based on conditioning a deep generative adversarial network (GAN) to a latent regressor using an unsupervised DL-based DIR network to generate realistic 4D-XCAT phantoms for simulation studies. Romaguera \emph{et al.} \cite{romaguera_probabilistic_2021} use similar concepts to generate 4D MRI volumes. As these methods use unsupervised DL-based DIR, they do not require to know the actual spatial transformation to be learned between the phases. Instead, similarly to conventional DIR methods that explicitly optimize a parameterized transformation \cite{vishnevskiy_isotropic_2017,heinrich_mrf-based_2013,jaouen_regularized_2021}, unsupervised DL-based DIR methods rely on maximizing (differentiable) similarity metrics while constraining DVF smoothness. A major argument in favor of unsupervised learning is that it is impossible to obtain real ground truth DVFs, and that supervised training on synthetic transformations through random warping \cite{eppenhof_pulmonary_2019, hoffmann_synthmorph_2022} can lead to unsatisfactory results \cite{fu_deep_2020}. Unsupervised DL-based DIR, however, has not yet reached the accuracy and robustness of conventional, non-learning-based DIR \cite{heinrich_voxelmorph_2022,hering_learn2reg_2022}. This is especially true beyond the cranial vault (e.g. in the lungs or the pelvis) where larger anatomical deformations and sharp intensity transitions may occur. As of today, no current state-of-the-art DL-based DIR methods \cite{hansen_graphregnet_2021, jiang_multi-scale_2020} can achieve substantial accuracy in 4DCT imaging, with a general tendency to produce overly smoothed DVFs, for instance, leading to unrealistic modeling of motion at sliding interfaces (Fig. \ref{dir}). On the other hand, conventional methods such as \cite{vishnevskiy_isotropic_2017} consistently achieve sub-millimeter average target registration errors (TRE) across public lung datasets even with large deformations \cite{huang_deep_2022, jaouen_regularized_2021}, which is better than current learning-based techniques \cite{hansen_graphregnet_2021, wang_plosl_2022, chi_few-shot_2022, wolterink_implicit_2022}. For this reason, in this paper, we propose to learn DVFs conditioned on the patient's respiratory amplitude in a supervised fashion leveraging state-of-the-art conventional DIR.  

In previous works, we have investigated the possibility of synthesizing respiratory phases from static 3D images by conditioning an image-to-image GAN architecture on 1D respiratory breathing traces \cite{jaouen_4d_2019,cao_image_2021}. The overall motion of the lung was reproduced, with a lowering of the diaphragm consistent with the corresponding scalar amplitude injected into the model. The approach was validated quantitatively using global metrics such as lung volume similarity and global image quality metrics (mean absolute error - MAE, root mean square error - RMSE and structural similarity - SSIM). However, the accuracy at smaller scales was not satisfactory (e.g. little to no tumor motion), which was not captured by the global evaluation metrics considered in the experiments. Critically, tumor motion was not accurately reproduced, especially in the lower lobes \cite{cao_patient-specific_2022}. This renders the approach of little practical value in RT treatment planning. More recently, similar global, image-level validation strategies were considered in two unsupervised learning-based approaches. Sang and Ruan performed deep interpolation between EOE and EOI phases by generating spatio-temporally smooth DVF \cite{sang_conditional_2023}. Lee \emph{et al.} \cite{lee_rmsim_2023} proposed a conditional seq2seq DVF learning architecture also using explicit smoothness penalty, requiring only one 3D CT image and a respiratory trace extracted from the position of the diaphragm. 
However, as previously mentioned, unsupervised learning based on smoothness constraints on the DVF may lead to poor realism, especially in the lungs \cite{vishnevskiy_isotropic_2017}, which is not captured by the global evaluation metrics reported in these works. It is thus critical for accurate tumor motion synthesis to further validate synthetic motion at a more local level, supporting major changes in the methodology and the validation presented in the current work.

\section{Methods and Materials}

Fig. \ref{flow} illustrates the proposed method's overall workflow. During training, the model learns a mapping between the EOE phase and the various phases of real 4DCT images. Generation is conditioned on the input phase and an external surface-based estimation $\alpha(t)$, which in clinical routine can be extracted from different clinically used respiratory monitoring devices (e.g. pneumatic belts, RPM, optical surface imaging). The generator maps to the image-$\alpha$ input pair a 3-component DVF in a supervised fashion based using a state-of-the-art conventional DIR method \cite{vishnevskiy_isotropic_2017}. During training, the output DVFs are compared to their ground truth following an $\ell_1$ minimization objective. A key contribution is to further improve transformation realism through an adversarial term based not only on the warped input, as is typically done in DL-based DIR, but also on the magnitude of the generated DVF itself. 

At test-time, given a 3D CT input image and corresponding $\alpha$ values, the model reproduces synthetic respiratory phase images that mimic breathing at the specified amplitudes. 

\subsection{Problem formulation and notations}

Let ${X}(\bm x)$ be a 3D CT image, where $\bm{x}\in \mathbb{R}^{3}$ is the image domain. Let also $\mathcal{Y}(\bm x)$ be a 4DCT image, i.e. a dynamic, vector-valued series of $N_g$ gates 
\begin{equation}
    \mathcal{Y}(\bm x) = \left({Y}_0(\bm x), ..., {Y}_k(\bm x),..., {Y}_{N_g}(\bm x) \right),
\end{equation} where $k\in \mathbb N$ is the gate number, with $k=0$ corresponding to the EOE phase without loss of generality. 

Our objective is to synthesize a vector-valued dynamic 4D image mimicking the breathing dynamics of a 4DCT acquisition from $X(\bm x)$, i.e. a set of $N$ 3D images :  
\begin{equation}
    \mathcal{Y}^*(\bm x) = \left(Y_0^*(X, \alpha_0),...,Y_k^*(X, \alpha_k), ..., Y_N^*(X, \alpha_N) \right),
\end{equation}
where $\alpha_k$ is the relative variation of respiratory amplitude (e.g. in mm) with respect to $X(\bm x)$. Doing so, we make the simplification that there is a one-to-one (bijective) correspondence between the true image $Y_k$ and respiratory amplitude $\alpha_k$, i.e. for each $\alpha_k$ there can only be one 3D CT image associated to it.

To this end, during training, we learn from a set of real 4DCT acquisitions $\left\{\mathcal Y^i\right\}$. For each training 4DCT image, we learn to reproduce a set of DVFs: 
\begin{equation}
    \left(\phi_{1}, ..., \phi_{k}, ..., \phi_{N_g} \right) : \mathbb{R}^3  \rightarrow \mathbb R^3,
\end{equation} between the EOE phase $Y_0(\bm x)$ and the remaining $N_g$ phases. To learn this transformation, we use a supervised DL-DIR network in the form of a U-Net shaped convolutional encoder-decoder with skip connections $G$, the last layer of which is a three-channel layer, one for each DVF dimension. To achieve amplitude-dependent image synthesis, $G$ is conditioned by a scalar value $\alpha_k$ that measures the relative respiratory amplitude between phases $Y_0$ and $Y_k$, i.e. $\phi^{*}_{\alpha_k}=G(Y_0, \alpha_{k})$ are the generated DVFs during training.

During deployment, DVFs are generated for any desired $\alpha$ value from a 3D CT image to obtain the requested image after warping $ Y^*(X, \alpha) = X \circ \phi^{*}_{\alpha}$, where ${X \circ \phi^{*}_{\alpha}}$ represents $X$ warped by $\phi^{*}_{\alpha}$. 

\subsection{Scalar conditioning layer}

To condition image generation to the respiratory pattern of the patient, we use an adaptive instance normalization (AdaIN) layer \cite{huang_arbitrary_2017} to adapt the features of $G$ to the respiratory amplitude $\alpha$ at the level of the bottleneck of the encoding branch. The original AdaIN layer applies the style of an image to a reference image in tasks such as style transfer by normalizing feature statistics using a variable channel-wise mean and variance. Here, the layer is modified to update the features based on $\alpha$ instead : 

\begin{equation}
   \textrm{AdaIN}(\bm{x},\alpha) = \gamma(\alpha)\left(\frac{\bm{x} - \mu(\bm{x})}{\sigma(\bm{x})}\right) + \beta(\alpha) ,
\end{equation}
where $\gamma$ and $\beta$ depend on $\alpha$ after passing through a fully connected layer and a one-dimensional convolutional layer. 
\subsection{DVF learning objective}

The training loss of our DVF synthesis model is a dual supervised-unsupervised compound loss :

\begin{equation}
    \mathcal{L}(G,D) = \lambda_{1}\mathcal{L}_{1}^{\phi_\alpha}(G)+ \lambda_{2} \mathcal L_\text{GAN}^{Y,||\phi_\alpha||}(G,D), 
\end{equation}
where $\lambda_1$ and $\lambda_2$ are parameters controlling the balance between these two terms.

Departing from unsupervised methods that would typically constrain DVF smoothness explicitly, we train $G$ in a supervised manner by adding a $\ell_1$ reconstruction loss on the marginal components of the DVF:
\begin{equation}
     \mathcal{L}_{1}^{\phi_\alpha} = \sum_{xyz}{||\phi^{*}_{\alpha_k} - \phi_k||_1}, 
\end{equation}
where each component of the field is treated independently. $\phi_k$ is the output of \textit{pTVreg} \cite{vishnevskiy_isotropic_2017}, arguably one of the best lung DIR algorithm based on isotropic total variation minimization, achieving consistently sub-millimeter accuracy across several public lung datasets \cite{huang_deep_2022,xiao_dual-supervised_2022}. 

To encourage the generation of realistic DVFs $\phi_\alpha^*$ without explicitly modeling field smoothness, we propose to train the model under a joint adversarial objective that concatenates both the warped image $Y_0 \circ \phi_\alpha^*$ and the magnitude of the DVF itself: 

\begin{equation} 
        \mathcal L_\text{GAN}^{Y,||\phi_\alpha||} = \mathbb{E}[\mathrm{log}D(X,Y_k,||\phi_k||)] + \mathbb{E}[\mathrm{log}(1-D(X, Y^*_k,||\phi^*_{\alpha_k}||)].
\end{equation}

The GAN loss evaluates the distribution alignment between the target image $Y_k$ and the generated image $Y^*_k$, i.e. the EOE phase $X$ warped by $\phi_k^*$. To further improve synthesis, the magnitude of the DVF is included in the adversarial cost. The rationale for this is that the information contained in the magnitude relates to the motion amplitude of the patient's respiratory pattern, which is associated to the scalar conditioning value $\alpha$. Higher values of $\alpha$ should thus correlate with higher magnitudes. We demonstrate in the experimental section that the addition of this term is indeed key to achieving more faithful motion synthesis. 

\begin{figure}[t]
    \centering

    \begin{subfigure}[t]{0.48\textwidth}
        \centering
        \includegraphics[width=\textwidth]{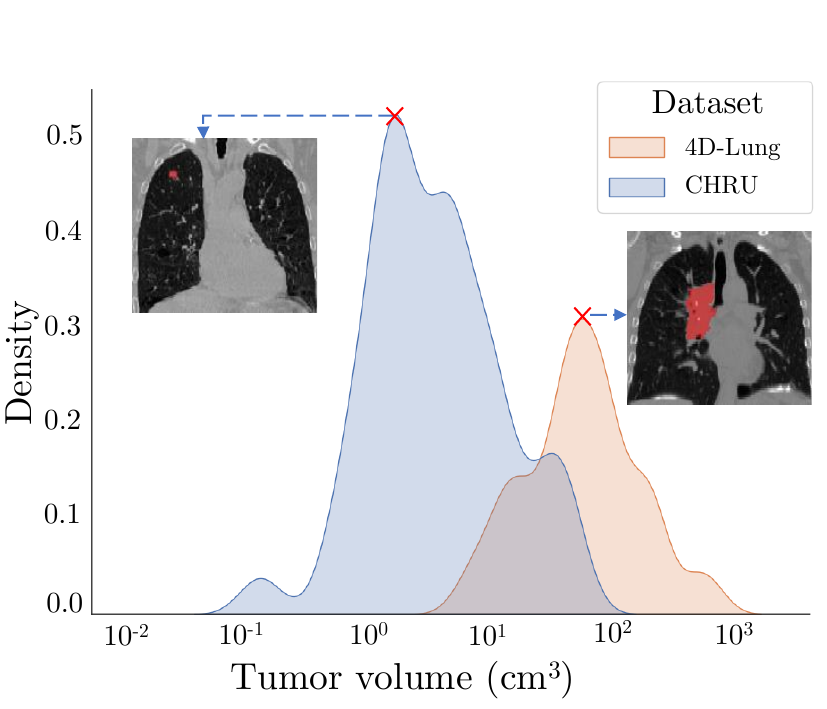}
        \caption{Distribution of tumor volumes in the 4D-Lung (red) and CHRU (blue) datasets.}
        \label{TumorsDistribution}
    \end{subfigure}
        \begin{subfigure}[t]{0.48\textwidth}
        \centering
        \includegraphics[width=\textwidth]{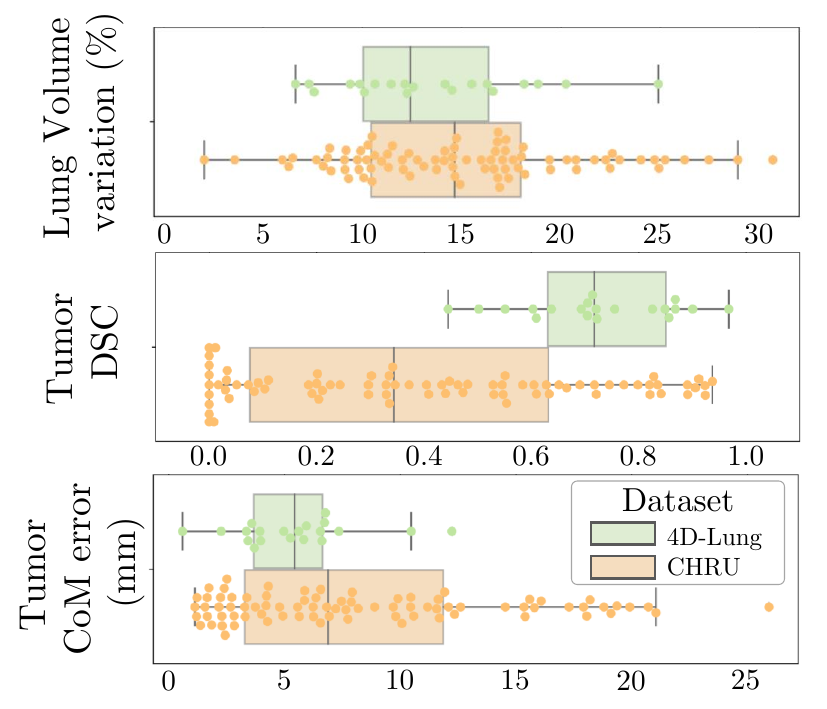}
        \caption{Metrics evaluated between EOE \& EOI phases}
        \label{DatasetVariation}
    \end{subfigure}
    \caption{Characteristics of the two 4DCT datasets used in our experiments. }
    \label{CHRUvsTCIA}
\end{figure}

\section{Validation setup}

\subsection{Quantitative metrics}

Different metrics were used for the validation of the proposed respiratory motion synthesis model.

\textbf{Reconstruction quality} We first considered global reconstruction metrics MAE and SSIM to evaluate the overall quality of the synthesized 4DCT phases $Y^*_k$ compared to the ground truth phase $Y_k$.

\textbf{Region-wise evaluation} We then measured the variation in global lung volumes by comparing both the synthetic and the true phase to the initial EOE phase $X$. The lung volume variation (VV) was defined as:

\begin{equation}
        \textrm{VV} = \frac{|V_B - V_A|}{V_A} \times 100, 
\end{equation}
where $V_A$ and $V_B$ are the lung volumes of the EOE phase and the target phase respectively. 

As already mentioned in the introduction, alongside global lung motion assessment, it is also crucial for RT treatment planning to evaluate the model accuracy at the tumor level, a problem somewhat overlooked by previous 4DCT synthesis reports \cite{sang_conditional_2023, lee_rmsim_2023}. To this end, we used the Dice Similarity Coefficient (DSC) between the GTVs delineated on $Y^*_k$ and the true phase $Y_k$. Considering the large variability of tumor volumes, we also used the Euclidean distance between the Centers of Mass (CoM) of the tumor as an additional volume-independent metric, hereafter referred to as the CoM error. 

Respiration affects not only the lungs but also peripheral organs at risk (OARs), the localization of which is paramount for accurate RT treatment planning. 
For this reason, we also evaluated volumetric DSC and CoM errors on a per-organ basis in the heart, liver, stomach, spleen, kidneys, and pancreas.

\textbf{DVF quality} To assess the quality of the generated DVFs, we quantified the total variation (TV) of the components of the DVF as in \cite{vishnevskiy_isotropic_2017}. The rationale is that well-behaved DVFs that are able to reproduce sharp transitions (e.g. at sliding interfaces in thoracic imaging) show lower TV than exceedingly smooth DVFs. Following common practice in DIR we also computed the negative determinant of the Jacobian of the transformation $|J(\phi^*)|$ to further characterize the smoothness of $\phi^*$. 

\subsection{Datasets}

We evaluated and validated the quality of our proposed dynamic image synthesis model using two 4DCT datasets that show remarkable differences. 

\textbf{4D-Lung dataset} We used a public dataset from The Cancer Imaging Archive (TCIA), hereafter referred to as \textit{4D-Lung} \cite{hugo_data_2016} and reported in other 4DCT synthesis studies \cite{huang_deep_2022, sang_conditional_2023}. It is composed of $20$ images of locally advanced, non-small cell lung cancer patients and includes GTV segmentation masks. Each 4DCT acquisition is composed of $10$ phases covering the full respiratory cycle. The images were acquired on a 16-slice helical CT scanner (Brilliance Big Bore, Philips Medical Systems) with a slice thickness of $3$mm and $512 \times 512$ axial resolution ($\sim1$mm pixel size). Tumors are of rather a large size with an average volume of $77\pm 96$ cm$^3$. This translates into a relatively high overlap in tumor volumes between the EOE and EOI phases, assessed by a DSC of $0.72\pm 0.14$ and tumor CoM errors of $5.46 \pm 2.66$ mm. 

\textbf{CHRU dataset} Metrics reported on tumors such as found in the 4D-Lung dataset hinder the fact that local accuracy may be insufficient to track smaller volumes, due e.g. to the well-known dependency of the DSC on object size \cite{reinke_common_2022}. For this reason, we also considered a local clinical database acquired at the radiation oncology department of Brest University hospital, France. This dataset referred to as the \textit{CHRU} dataset, contains patients with both primary tumors and lung metastases. Due to quality concerns with certain 4DCT acquisitions, not all data were included in the experiments. Images with acquisition issues, such as missing thoracic portions or significant motion artifacts, were excluded. Ultimately, only 43 patients were retained for the experiments. Compared to the 4D-Lung dataset, the CHRU dataset shows much smaller GTVs on average and greater volume variability, with mean volumes of $6.4\pm 9.1$ cm$^3$, which translates into much lower initial DSC $0.39\pm 0.31$ and greater distances between CoMs ($8.44 \pm 6.04$ mm). Another notable feature is inclusion of patients who underwent repeat 4DCT acquisitions, i.e. 30 patients had two 4DCT, five patients had three 4DCT, and one patient had four 4DCT. This enables the assessment of inter-4DCT variability, which is a key feature to evaluate the clinical interest of a respiratory motion synthesis model. Each CT acquisition was reconstructed into eight respiratory gates ($0\%$-to-$100\%$ inspiration followed by $75\%$-to-$25\%$ expiration by steps of $25\%$). The images were acquired with a Siemens Sensation Open, $2$mm slice thickness, and $512 \times 512$ axial resolution ($\sim1$mm pixel size).

Fig. \ref{TumorsDistribution} shows the distribution of tumor volumes in the 4D-Lung and CHRU datasets, with two typical examples. To further characterize these differences, Fig. \ref{DatasetVariation} shows the distribution of distances between the EOE and EOI phases in each 4DCT acquisition, which relates to the amplitude of motion correction required by the motion synthesis approach. Compared to 4D-Lung, the CHRU dataset shows greater volumes differences between EOE and EOI phases, smaller GTVs, and a greater diversity of tumor appearances.

\begin{figure}[!]
    \centering
    \begin{subfigure}[t]{0.24\textwidth}
        \centering
        \captionsetup{justification=centering}
        \includegraphics[width=\textwidth]{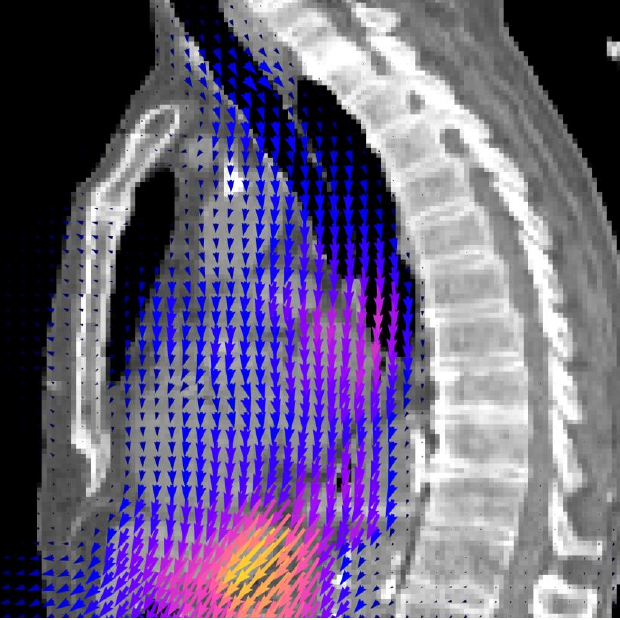}
        \caption{Reconstruction loss only ($\mathcal L_1^{\phi}$)}
        \label{noGAN}
    \end{subfigure}
    \begin{subfigure}[t]{0.24\textwidth}
        \centering
        \captionsetup{justification=centering}
        \includegraphics[width=\textwidth]{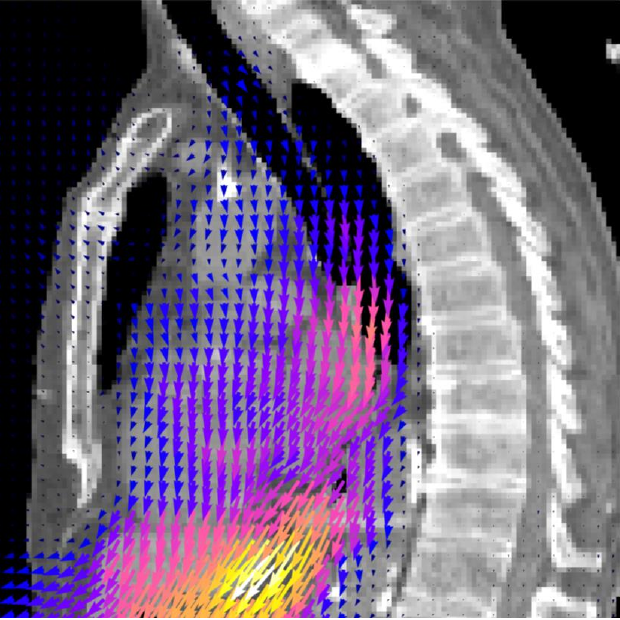}
        \caption{w/ unsupervised GAN loss ($\mathcal L_1^{\phi}+\mathcal L_\text{GAN}^{Y}$)}
        \label{GAN_image}
    \end{subfigure}
    \begin{subfigure}[t]{0.24\textwidth}
        \centering
        \captionsetup{justification=centering}
        \includegraphics[width=\textwidth]{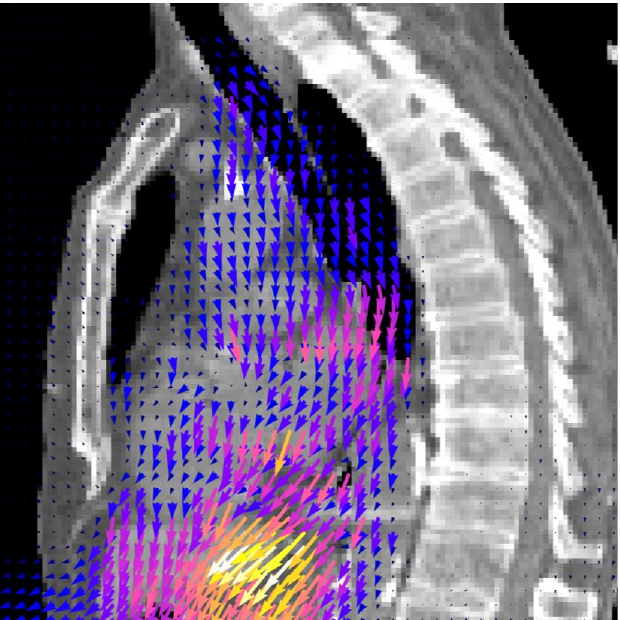}
        \caption{Proposed joint loss ($\mathcal L_1^{\phi}+\mathcal L_\text{GAN}^{Y,||\phi_{\alpha}||}$)}
        \label{GAN_mag}
    \end{subfigure}
    \begin{subfigure}[t]{0.24\textwidth}
        \centering
        \captionsetup{justification=centering}
        \includegraphics[width=\textwidth]{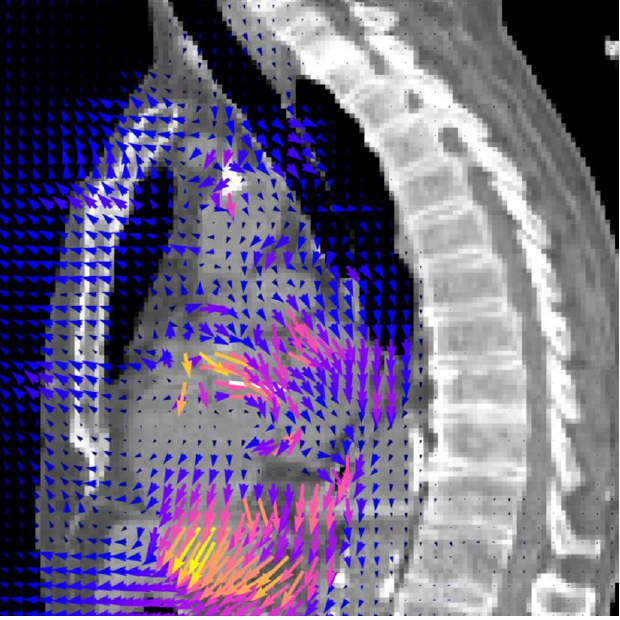}
        \caption{Reference}
        \label{pTVreg}
    \end{subfigure}
    \caption{Influence of the loss terms on the generated DVF. (a) Reconstruction loss only: $\ell_1$ on the DVF (b) Addition of an adversarial term on the warped image (c) Proposed compound loss: $\ell_1$ on the DVF + adversarial term jointly on the warped image and the DVF magnitude. (d) Reference state-of-the-art DVF obtained by \cite{vishnevskiy_isotropic_2017}}
    \label{comparisonDVF} 
\end{figure}

\subsection{Other experiments}

\textbf{4DCT variability study} Due to the intrinsic variability of respiratory patterns in patients acquired at different times, which relates to the inherent limitations of 4DCT treatment planning, a useful metric to validate the clinical relevance of the proposed respiratory motion synthesis method is to ascertain that the distance between the synthetic phase $Y^*_k$ and real phase $Y_k$ is not greater than the distance between repeat 4DCT acquisitions $Y_k^1$ and $Y_k^2$ of the same patient. Repeat scans were thus compared using DSC and CoM error to evaluate this variability.

\textbf{Compound loss validation} We quantified the improvement obtained by the proposed compound loss based on supervised learning combined to an adversarial loss on the magnitude of the DVF. To this end, we also trained alternative models in which the total loss included combinations of reconstruction and adversarial terms based on the warped image and/or the DVF. 

\textbf{Comparison to state of the art} We compared our approach to a recent work, RMSim \cite{lee_rmsim_2023}, that also generates 4DCT sequences from static images and respiratory traces. RMSim uses a modified seq2seq architecture based on unsupervised DVF learning to enforce temporal consistency of the motion. Contrary to our approach relying on external respiratory measurements, the amplitude was obtained from image-derived internal information (maximum displacement of the diaphragm in the superior-inferior axis), which is not realistic and cannot be obtained in practice. The inference code was made available by the authors of RMSim at the time of the completion of this work. To ensure a fair comparison, we retrained our approach using similar image-derived respiratory guidance, by measuring superior-inferior motion of the diaphragm, since our method is adaptable to any form of scalar conditioning. We did not retrain RMSim and used the network weights of the authors. We also used the same training conditions for our method and learned a mapping from EOI to EOE phase, rather than from EOE to EOI. The authors trained their model on a private dataset using 100 patients with 10-phase 4DCTs using four A40 GPUs with 48GB memory each during 72 hours, which is considerably more than our proposed model (see next section). The inference stage of RMSim required 35.2 GB of GPU memory. We compared the performances of these two models on $20$ patients using the public 4D-Lung dataset.   % they trained on a private dataset, we did the same, and for "equality", we tested on the same public dataset (un truc de ce genre). 
% RMSim : The model was trained and tested on an internal high performance computing cluster with 4 NVIDIA A40 GPUs with 48GB memory each. Our model consumed 35.2 GB GPU memory and the training time was approximately 72 hours. The inference time for 9 phases and 40 total test cases from the internal dataset was less than 3 minutes.

\subsection{Implementation and training} 

The model was trained in 3D using $42$ 4DCT images from $20$ patients from the CHRU dataset. The testing set was composed by another $44$ 4DCT images from $23$ additional patients of the CHRU dataset and $20$ 4DCT images from $20$ patients in the 4D-Lung dataset, which are also used for model comparison with RMSim. The model was trained for $200$ epochs using ADAM optimization \cite{kingma_adam_2017}.
Loss hyperparameters were set to $\lambda_1=100$ and $\lambda_2=1$. 

The initial 4DCT from the dataset were not aligned and had varying sizes in the superior-inferior (SI) axis. We manually selected a reference 4DCT scan with a suitable FOV along the SI axis and resampled all other 4DCT scans by aligning the centers of origins. Images were then cropped around the lungs using a robust DL-based automatic lung segmentation method \cite{hofmanninger_automatic_2020}. We then selected the largest bounding box (EOI phase), which we extended by $2$cm past the lower point of the lungs to include abdominal organs of interest. Finally, each image was resampled to a size of $128 \times 128 \times 128$ to allow the neural network to be trained with relatively modest GPU capability. In all our experiments, we used an NVIDIA GeForce GTX 1080Ti graphics card with 11GB of memory and a batch size of 1. 

Tumor masks available for each phase of the 4D-Lung dataset were used in experiments requiring GTV delineations. For the CHRU dataset, GTVs were obtained automatically in all phases using a nnU-Net model \cite{isensee_nnu-net_2021} trained with the winning weights of the lung tumor task of the Medical Segmentation Decathlon \cite{antonelli_medical_2022}. Heart segmentations were performed using PlatiPy, a free RT segmentation software leveraging nnU-Net with additional image processing heuristics \cite{finnegan_feasibility_2019}. Remaining organ segmentations (liver, stomach, kidneys, and spleen) were obtained by training a nnU-Net model on the Multi-Modality Abdominal Multi-Organ Segmentation Challenge (AMOS) 2022 training dataset \cite{ji_amos_2022}. In all cases, final segmentation masks were reviewed by an expert and manually corrected when necessary. 

For each available repeat scan in the 4DCT variability study, we performed a rigid registration between the EOE phases to avoid offsets in tumor and organ positions. The CoM error was then computed between the two tumor segmentations. The distance error in each axis was used to register the second scan, correcting the initial CoM error. Subsequently, the transformation was applied to the other phases and the segmentations of the repeat scan to ensure consistency.%\footnote{Details about corrections are available in Supplementary Materials.}.

The scalar conditioning value $\alpha$ was computed from respiratory amplitude measurements in a fully automated fashion. For each 4DCT phase of the two datasets, we computed a respiratory amplitude (in mm) from the image following the method described in \cite{fayad_technical_2011}, thus mimicking the knowledge that can be obtained from a respiratory tracking belt such as the Anzai Respiratory Gating System. Each phase was segmented to obtain the patient's body \cite{zhou_automated_2004}. Binary difference images were obtained between the EOE and the other phases. The mean distance in the anterior-posterior axis was defined as $\alpha$. 

\section{Results}

%Fig. \ref{transition} shows synthetic phases obtained by our generator from an EOE phase of a patient for varying values of $\alpha$, which were compared against the real corresponding 4DCT acquisition. Visually, a lowering of the diaphragm was achieved with increasing values of $\alpha$, in good agreement with the real 4DCT phases. One can note that the model is also capable of interpolation (i.e. generating images between two real respiratory phases), and of extrapolation (i.e. lowering the maximum diaphragm position beyond the true EOI phase). 

\subsection{Overall image and DVF analysis}

To illustrate qualitatively the effect of the proposed compound loss used in the training of our model, Fig. \ref{comparisonDVF} compares generated DVFs obtained using other variants of the total loss. When using only the reconstruction loss on the DVF (Fig. \ref{noGAN}), the generated DVF was smoother and largely unidirectional along the superior-inferior direction, with a low absolute magnitude corresponding to a weaker lowering of the diaphragm. When an adversarial term was added only considering the warped image $Y^*=X\circ \phi^*_k$, as can be done for instance in unsupervised DL-DIR \cite{fu_lungregnet_2020}, the magnitude was closer to the reference but nevertheless remained exceedingly smooth (Fig. \ref{GAN_image}). More faithful DVF appearance was achieved through the proposed dual supervised-unsupervised loss taking also into account the magnitude of the DVF $||\phi^*||$ in the joint adversarial cost (Fig. \ref{GAN_mag}). Sharper transitions and orientations were achieved, in better agreement with the DVF obtained with the reference registration method pTVreg (Fig. \ref{pTVreg}).

To support these visual observations, Table \ref{model_comparison} shows results obtained using other combinations of loss terms. Training using only reconstruction losses in the supervised setting $(\mathcal L_1^{\phi})$ or as in unsupervised DL-DIR using the warped image $(\mathcal L_1^{Y})$ led to relatively high reconstruction errors (mean MAE $\simeq$ 55 and mean SSIM $\simeq$ 0.57). When the $\ell_1$ loss was combined to a GAN loss on the warped image $Y^*$ only (such as in \cite{fu_lungregnet_2020} for DL-DIR), excessive DVF smoothness could be captured by high TV and Jacobian values. On the other hand, best quantitative scores with respect to all metrics considered were achieved when the adversarial term jointly considered the magnitude of $\phi^*$, with much higher average SSIM (0.74$\pm$0.05) and lower MAE (31.75$\pm$5.23), as well as smoothness measures comparable to the reference DVF obtained through pTVreg.

\begin{table}[t]
    \centering
    \caption{Compound loss analysis.}
    \label{model_comparison} 
   % \resizebox{\columnwidth}{!}{
    \begin{tabular}{@{}lcccc@{}}
    \toprule
                                                    & \multicolumn{2}{c}{Warped image}                             & \multicolumn{2}{c}{DVF}   \\ \cmidrule(l){2-5} 
                                                    & \multicolumn{1}{c}{MAE}    & \multicolumn{1}{c}{SSIM} & \multicolumn{1}{c}{TV} & \multicolumn{1}{c}{$|J|$}   \\ \midrule
     No synthesis                                  &      54.53$\pm$8.98       & 0.53$\pm$0.06           & N/A    & N/A\\\midrule

    $\mathcal L_1^{Y}$                                                                  & 55.77$\pm$13.17                & 0.57$\pm$0.10                 & 0.25$\pm$0.08    & 1.037$\pm$0.018\vspace{.1cm}\\
    $\mathcal L_1^{\phi}$                                                               & 56.29$\pm$13.79                & 0.57$\pm$0.10                 & 0.26$\pm$0.08    & 1.037$\pm$0.016\vspace{.1cm}\\
    $\mathcal L_1^{Y}+\mathcal L_\text{GAN}^{Y}$                                        & 43.00$\pm$10.64                & 0.57$\pm$0.07                 & 0.41$\pm$0.08    & 1.028$\pm$0.010\vspace{.1cm}\\
    $\mathcal L_\text{GAN}^{Y,||\phi_{\alpha}||}$                                       & 42.59$\pm$13.85                & 0.63$\pm$0.10                 & 0.21$\pm$0.02    & 1.004$\pm$0.002\vspace{.1cm}\\
    $\mathcal L_1^{\phi}+\mathcal L_\text{GAN}^{Y}$                                     & 37.39$\pm$12.12                & 0.68$\pm$0.10                 & 0.12$\pm$0.04    & 1.008$\pm$0.005\vspace{.1cm}\\
    $\mathcal L_1^{\phi}+\mathcal L_\text{GAN}^{Y,||\phi_{\alpha}||}$                            & \textbf{31.75$\pm$5.23}                 & \textbf{0.74$\pm$0.05}                 & \textbf{0.12$\pm$0.04}    & \textbf{1.006$\pm$0.004}\vspace{.1cm}\\
    \midrule
    Reference DVF                                                            &      N/A       & N/A           & 0.13$\pm$0.03    & 1.007$\pm$0.006\\ \bottomrule

    \end{tabular}
   % }
\end{table}

Table \ref{global_table} summarizes reconstruction results assessed by quantitative metrics MAE and lung volume variation (VV). In the 4D-Lung dataset, mean VV was $6.86\%$ and mean MAE was $35.71$ before motion synthesis, suggesting relatively moderate volume changes between the EOE and the other phases. After motion synthesis, VV was reduced to $1.50\%$ on average. In the CHRU dataset, higher initial lung VV of $11.66\%$ was reduced down to $3.05\%$ on average after motion synthesis. MAE was also improved substantially. Training our model on CHRU only was not detrimental to the performances achieved on the 4D-Lung dataset, suggesting good generalization ability even when datasets have quite different properties.

\begin{table}[hbtp]
    \centering
    \caption{Overall evaluation analysis.}
    %\resizebox{\columnwidth}{!}{
    \begin{tabular}{@{}lll@{}}
    \toprule
                                   & 4D-Lung  &  CHRU     \\ \midrule
    Initial VV (\%)                 & 6.86 $\pm$ 5.32   & 11.66 $\pm$ 5.13  \\
    After synthesis VV (\%)         & 1.50 $\pm$ 1.19   & 3.05 $\pm$ 2.26   \\
    Initial MAE                    & 35.71 $\pm$ 15.95 & 56.47 $\pm$ 10.45 \\
    After synthesis MAE            & 29.87 $\pm$ 10.74 & 42.26 $\pm$ 6.35  \\ \bottomrule
    \end{tabular}
    %}
    \label{global_table}
\end{table}

\subsection{Region-wise evaluation}

\begin{figure}[b]
    \centering
    \begin{subfigure}[b]{0.32\columnwidth}
        \centering
        \includegraphics[width=\textwidth]{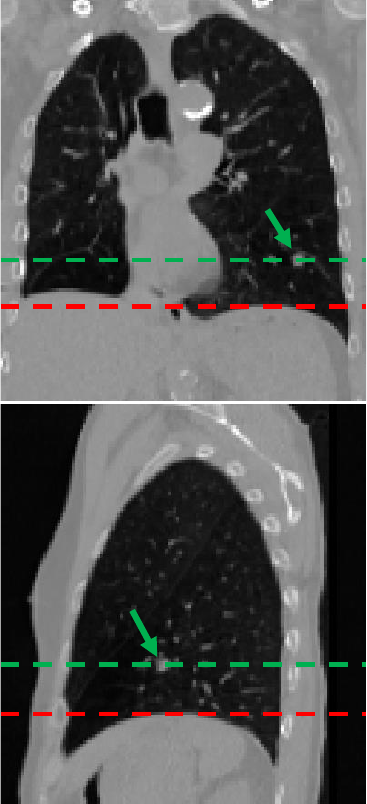}
        \caption{EOE phase}
        \label{comparaison_input}
    \end{subfigure}
    \begin{subfigure}[b]{0.32\columnwidth}
        \centering
        \includegraphics[width=\textwidth]{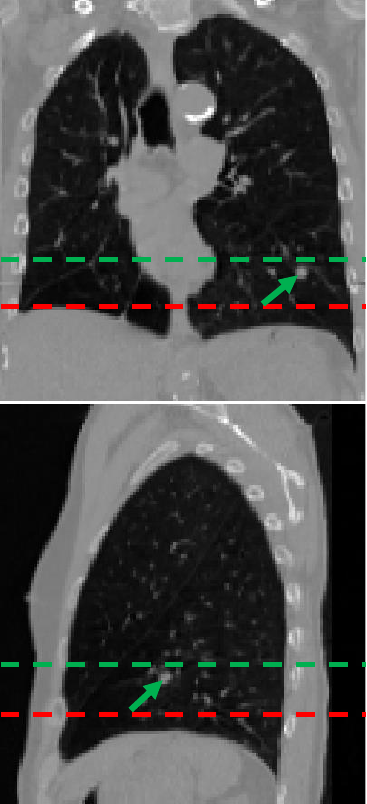}
        \caption{Ground truth}
        \label{comparaison_target}
    \end{subfigure}
    \begin{subfigure}[b]{0.32\columnwidth}
        \centering
        \includegraphics[width=\textwidth]{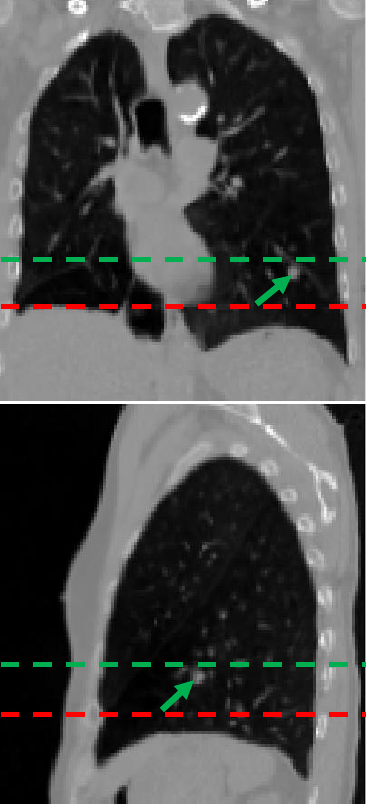}
        \caption{Ours}
        \label{comparaison_ours}
    \end{subfigure}
    \caption{Global lung motion on the coronal and sagittal plane. The red and green lines represent the initial position of the diaphragm in the left lung and the tumor respectively. The green arrow shows the location of the tumor.}
    \label{comparison} 
\end{figure}

\begin{table}[t]
    \centering
    \caption{Inter-phase similarity metrics at the tumor level before (no motion synthesis) and after synthesis.}
    %\resizebox{\columnwidth}{!}{
    \begin{tabular}{@{}lll@{}}
    \toprule
                                   & 4D-Lung  &  CHRU     \\ \midrule
    Initial DSC                    & 0.85 $\pm$ 0.13   & 0.43 $\pm$ 0.30   \\
    After synthesis DSC            & 0.89 $\pm$ 0.09   & 0.68 $\pm$ 0.22   \\
    Initial CoM error (mm)         & 2.83 $\pm$ 2.49   & 6.93 $\pm$ 4.97   \\
    After synthesis CoM error (mm) & 1.52 $\pm$ 1.29   & 2.74 $\pm$ 1.69   \\ \bottomrule
    \end{tabular}
    %}
    \label{local_table}
\end{table}

Fig. \ref{comparison} shows a synthetic 4DCT phase from a patient of the CHRU dataset with a relatively small tumor ($0.54$ cm$^3$) visible both in the sagittal and coronal planes. The tumor and the diaphragm position after motion synthesis matched well the ground truth image in both planes (Fig. \ref{comparaison_target}). \footnote{Animated visuals are available in Supplementary Files.}%\footnote{Another example with a larger tumor ($31.56$ cm$^3$) is shown in Supplementary Materials.} % Difference images between phases before (Fig. \ref{tumorMotion}b) and after motion synthesis (Fig. \ref{tumorMotion}c) illustrate the ability of the model at reproducing tumor motion faithfully.

\begin{figure}[b]
    \centering
    
    \begin{subfigure}[b]{\columnwidth}
        \centering
        \includegraphics[width=0.875\textwidth]{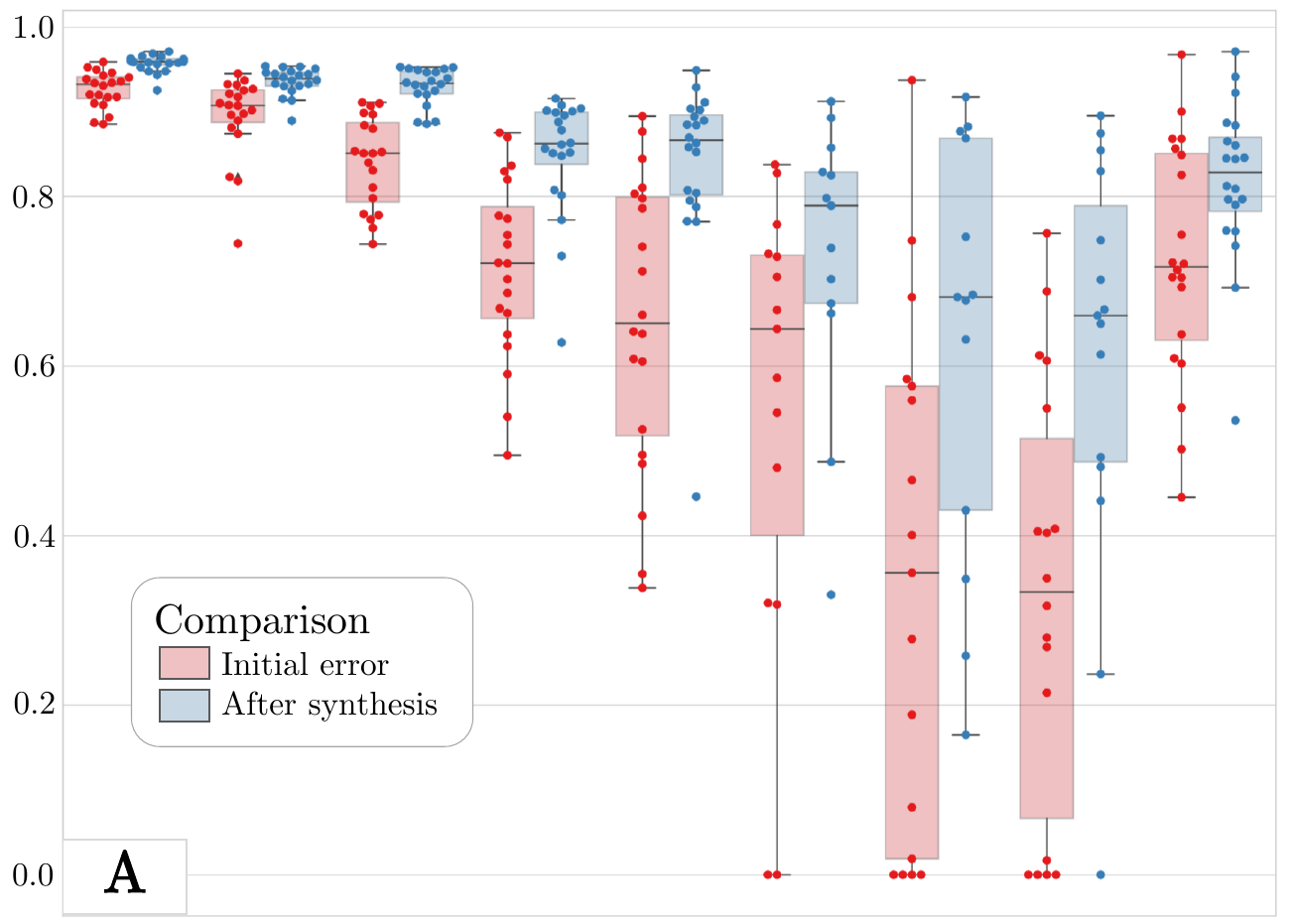}
    \end{subfigure}
    \begin{subfigure}[b]{\columnwidth}
        \centering
        \includegraphics[width=0.875\textwidth]{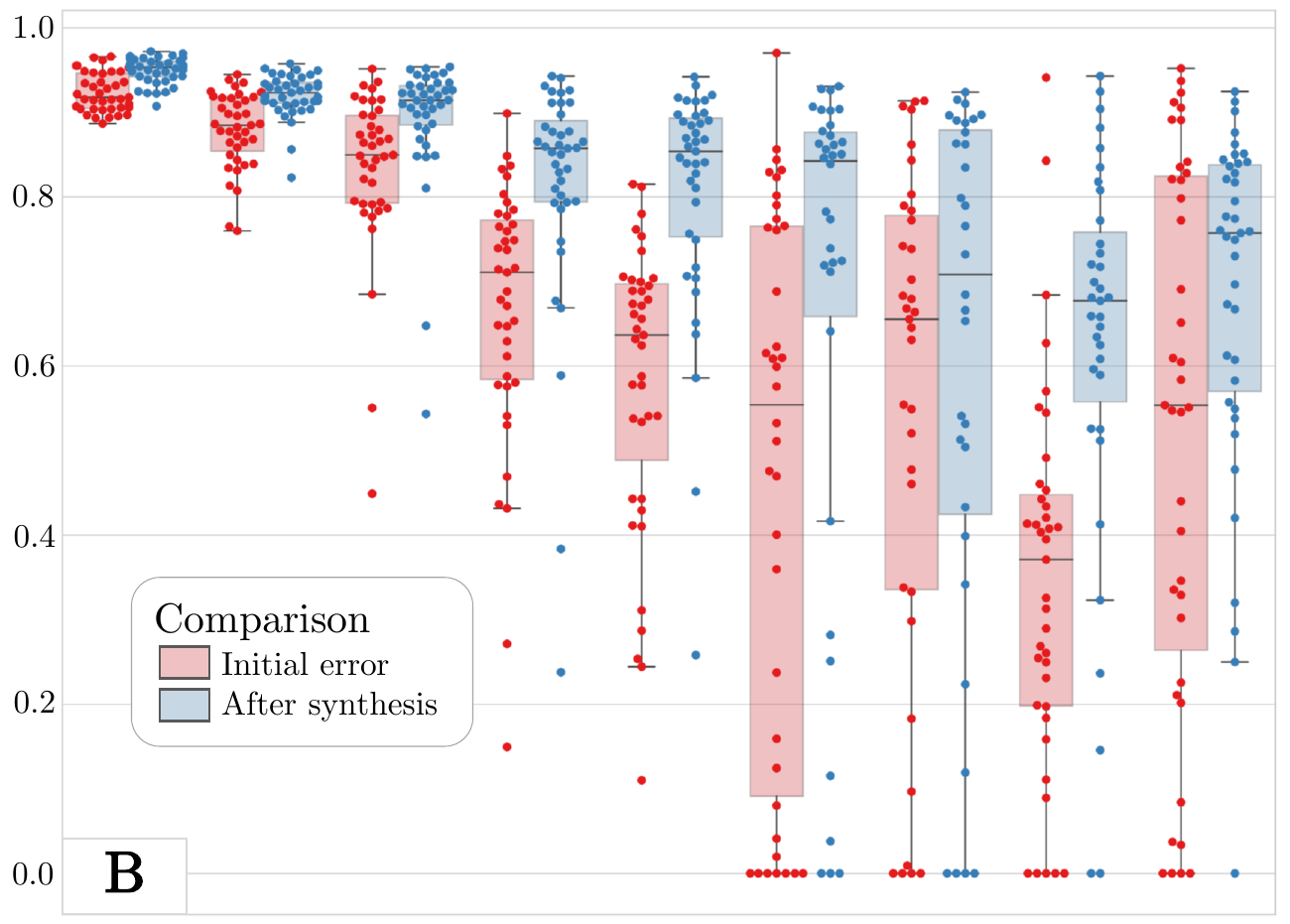}
    \end{subfigure}
    \hspace*{3mm}%
    \begin{subfigure}[b]{0.095\columnwidth}
        \caption{}
    \end{subfigure}
    \begin{subfigure}[b]{0.095\columnwidth}
        \caption{}
    \end{subfigure}
    \begin{subfigure}[b]{0.095\columnwidth}
        \caption{}
    \end{subfigure}
    \begin{subfigure}[b]{0.095\columnwidth}
        \caption{}
    \end{subfigure}
    \begin{subfigure}[b]{0.095\columnwidth}
        \caption{}
    \end{subfigure}
    \begin{subfigure}[b]{0.095\columnwidth}
        \caption{}
    \end{subfigure}
    \begin{subfigure}[b]{0.095\columnwidth}
        \caption{}
    \end{subfigure}
    \begin{subfigure}[b]{0.095\columnwidth}
        \caption{}
    \end{subfigure}
    \begin{subfigure}[b]{0.095\columnwidth}
        \caption{}
    \end{subfigure}
    \caption{Comparison of organs overlap between images for the EOI phase in 4D-Lung dataset (A)  and CHRU dataset (B). (a) Lung (b) Heart (c) Liver (d) Stomach (e) Spleen (f) Left Kidney (g) Right Kidney (h) Pancreas (i) Tumor.}
    \label{organDice} 
\end{figure}

Fig. \ref{organDice} shows box-plots of volumetric overlaps between EOI and EOE phases for the organs considered in our experiments in both datasets. For all organs, DSC was substantially improved after motion synthesis. This indicates that our model is not only capable of modeling motion inside the lungs but can also deform peripheral organs with good accuracy, a higher overlap after synthesis in all organs (DSC $\geq 0.65$). 

Table \ref{local_table} shows motion synthesis results at the tumor level in both datasets. As previously discussed, the 4D-Lung dataset showed lower initial CoM errors between tumors with a mean of $2.83$mm and a mean tumor DSC of $0.85$ due to the fact that tumors are of rather large size and motion relatively. In contrast, the CHRU dataset had larger mean initial CoM errors ($6.93$mm) and lower mean initial DSC ($0.43$). After synthesis, the mean CoM error was reduced to $1.52$mm for a mean DSC of $0.89$ in the 4D-lung dataset (resp. mean CoM error of $2.74$mm and mean DSC of $0.68$ in the CHRU dataset). Despite differences in tumor characteristics, the model was thus able to displace tumors in both datasets in a realistic and accurate fashion.

\begin{figure}[t]
    \centering
    \includegraphics[width=0.85\columnwidth]{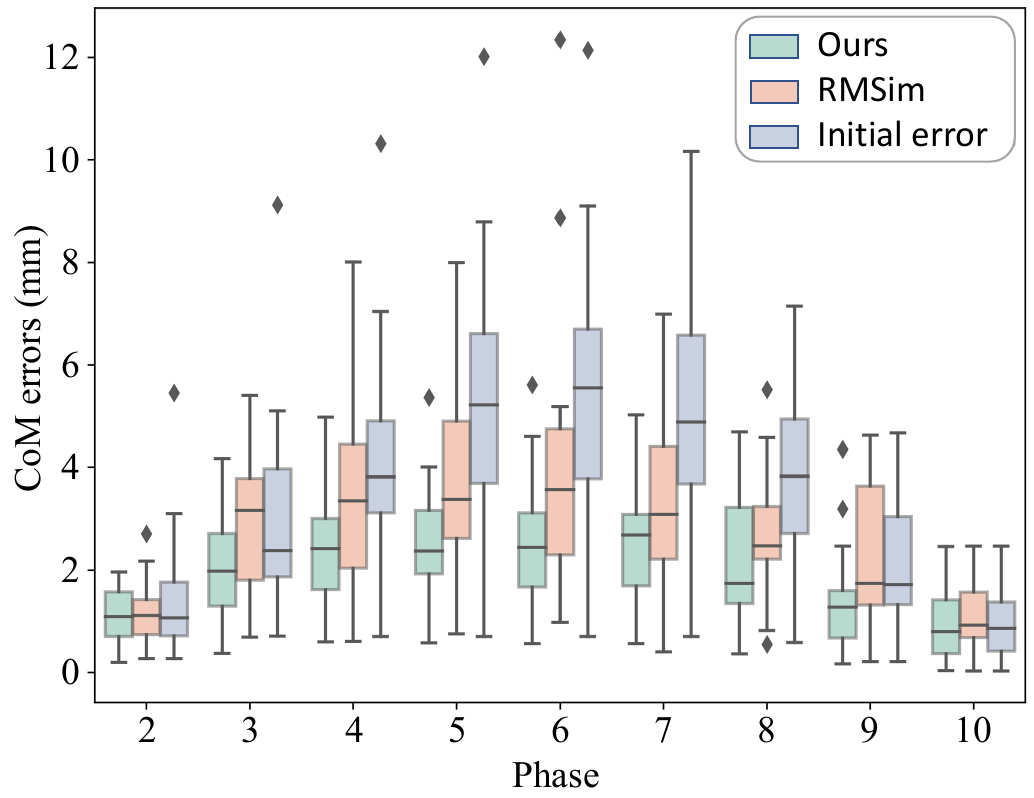}
    \caption{Tumor motion synthesis accuracy along the respiratory phases in the 4DCT-Lung dataset (assessed by CoM errors in mm). Our approach is compared to RMSim, a recent work based on unsupervised learning of DVF fields with temporal regularization \cite{lee_rmsim_2023}.}
    \label{fig:com_rmsim}
\end{figure}

% The approach is a modified seq2seq that captures spatial-temporal respiration patterns signal thanks stacked 3D-ConvLSTMs. It allows to predict future breathing phases given 3D CT image and auxiliary 1D respiratory trace.  % On en parle déjà rapidement dans l'intro ? Besoin de compléter ?
% To ensure a fair comparison, we developed an alternative model using identical settings to theirs. The model was subsequently trained with the goal of generating the EOI phase from the EOE phase, rather than the EOE phase to EOI phase, as initially designed in our method. Additionally, 
%To ensure a fair comparison to RMSim, We conditioned our model with the maximum displacement in the superior-inferior axis of the diaphragm, as described in their method. It is important to note that this conditioning does not align with clinical routine practices. 

\begin{table*}[hbtp]
    \centering
    \caption{4DCT variability study results}
    \resizebox{\textwidth}{!}{
     \begin{tabular}{@{}ll|llllllll@{}}
 \toprule
                   & \multicolumn{1}{c}{CoM error (mm)} & \multicolumn{8}{c}{DSC}                                                                           \\ \midrule
                   & Tumor                    & Tumor        & Heart         & Liver         & Stomach        & Spleen       & Right Kidney     & Left Kidney   & Pancreas  \\ \midrule
 4DCT1 vs 4DCT2    & 1.97$\pm$1.05         & 0.63$\pm$0.25 & 0.90$\pm$0.04 & 0.88$\pm$0.05 & 0.67$\pm$0.13 & 0.70$\pm$0.09 & 0.69$\pm$0.14    & 0.68$\pm$0.20   & 0.58$\pm$0.10 \\
 Ours1 vs 4DCT1    & 2.35$\pm$0.08         & 0.71$\pm$0.15 & 0.91$\pm$0.02 & 0.91$\pm$0.03 & 0.78$\pm$0.05 & 0.78$\pm$0.08 & 0.72$\pm$0.15    & 0.67$\pm$0.15   & 0.69$\pm$0.11 \\ 
 Ours2 vs 4DCT2    & 2.86$\pm$1.32         & 0.67$\pm$0.25 & 0.93$\pm$0.02 & 0.90$\pm$0.04 & 0.79$\pm$0.07 & 0.78$\pm$0.07 & 0.67$\pm$0.16    & 0.69$\pm$0.17   & 0.63$\pm$0.15 \\ \bottomrule
    \end{tabular}
    }
    \label{interscan} 
\end{table*}

Table \ref{interscan} shows quantitative results corresponding to the 4DCT variability study, where we evaluated the agreement between two acquired 4DCT scans compared to the agreement between real and synthetic scans. Within this context, the EOI phase of the first 4DCT acquisition (4DCT1) was compared against the EOI phase of the second scan (4DCT2). Mean inter-scans tumor CoM errors and DSC were $1.97$mm and $0.63$ respectively, stressing the inherent variability in the observed tumor position between real repeat 4DCT acquisitions. Comparing real and synthetic phases led to similar mean CoM errors of $2.35$mm and $2.86$mm and mean DSC of $0.71$ and $0.67$ for the tumor. Similar results were achieved regarding volumetric overlaps between OARs, with a variability comparable between images of acquired repeat and synthetic images. % Peut etre mieux de mettre en discussion ? En rajoutant l'importance d'utiliser des métriques appropriées pour mieux évaluer voire éviter des limitations de certaines métriques ? Ca fait plus interpretation que 'analyse' simple de résultat non 

Fig. \ref{fig:com_rmsim} compares tumor synthesis accuracy in the 4D-Lung dataset between our approach and RMSim \cite{lee_rmsim_2023}. Initial CoM errors for the EOI phase were $5.46$ mm on average, suggesting moderate tumor motion. Our model consistently outperformed RMSim across all respiratory phases, despite a smaller training dataset and lower computational requirements during both training and testing. %\footnote{DSC results for OARs are available in the Supplementary Materials.}.

\section{Discussion}

Our main contribution lies in training the proposed model using supervised learning on highly realistic DVF labels while using an adversarial term jointly not only on the warped image (i.e. similar to unsupervised DL-DIR training) but also critically on the magnitude of the field itself, a method that combines advantages of both supervised and unsupervised learning. The motivation for this strategy is to improve DVF realism and to indirectly encourage correlations with the respiratory amplitude that is passed simultaneously to the model, without constraining the smoothness of the DVF through explicit regularization (e.g. as in RMSim \cite{lee_rmsim_2023}). Our results demonstrate that this indeed substantially improves reconstruction quality compared to conventional reconstruction losses or when using an adversarial term on the warped image only, such as is typically done in unsupervised DL-DIR \cite{fu_lungregnet_2020}. This result, although being original in medical imaging, agrees with earlier observations that adversarial losses are more effective for image-to-image translation when reconstruction is not overly constrained \cite{groenendijk_benefit_2020}. This is highly relevant to our application and likely explains why our prior efforts using direct reconstruction losses did not produce satisfactory results at small scales \cite{cao_image_2021, cao_patient-specific_2022}.

In RT, accurate estimation of the motion of the tumors and the neighboring OARs is critical for calculating the dose delivered to the patient and thus maximizing treatment efficacy and minimizing the risk of radiation-induced side effects. Using only the output of necessarily imperfect DVFs to evaluate model performance is not satisfying. For this reason, evaluation was performed using ground truth 4DCT acquisitions. The method was extensively validated using two datasets that show notable differences in terms of GTV distributions. We evaluated the robustness of our results at multiple scales considering global image fidelity, global lung motion, tumor motion, and also motion of neighboring OARs. Unlike previous studies that only considered global reconstruction metrics \cite{sang_conditional_2023, lee_rmsim_2023}, covering more aspects of the model's performance across different metrics and datasets allowed for a more comprehensive evaluation of the properties of the model. To remove dependency on object size, we calculated in addition to volumetric overlaps the Euclidean distance between the CoM of tumors as a volume-independent metric. Our results show that it is critical to not only rely on single volume-based metrics in order to report tumor tracking quality more faithfully \cite{huang_deep_2022}. The reported DSC values were lower between repeat 4DCT phases compared to errors synthetic images, indicating that our the position of the tumors were closer. The COM error focuses primarily on the tumor's overall position, while the DSC coefficient considers both the shape and size of the tumor. In the case of repeated 4DCT scans, the tumor may undergo minor positional changes without significantly altering its overall shape. This results in a relatively low COM error, suggesting that the tumor's motion is similar between scans. However, the DSC coefficient may be lower if the tumor undergoes minor changes in shape or size between scans. Our model's predictions tend to closely resemble the shape of the tumor in the initial 3DCT scan, which is then used to predict subsequent phases. This similarity in shape between the initial 3DCT and synthetic phases facilitates better overlap between the segmentations of generated images and real 4DCT images, despite the slight positional discrepancies attributed to the COM error.

Despite leveraging the respiratory signal and image to enhance the accuracy of motion prediction, determining the actual tumor position remains a challenge due to the variation of the breathing patterns and other factors that may influence the tumor motion. 
A key result of our work is the demonstration that the error associated with motion synthesis is comparable with mismatches typically obtained between repeat 4DCT acquisitions of the same patient, which relates to the inherent limitations of 4DCT-based treatment planning. The average tumor CoM errors achieved were less than $3$mm, which is in line with a recent study of \cite{tremolieres_lung_2022} recommending PTV margins due to respiratory motion between $3$mm and $5$mm depending on tumor localization. Our results can thus be considered satisfactory in regard to the intrinsic variability of 4DCT treatment planning.

% Supervised learning limitations
While we used state-of-the-art DIR to supervise our model, we acknowledge that the inaccuracies in the DVFs may lead to errors in image synthesis. This is a typical problem in machine learning when using noisy labels for supervision. However, we showed that these errors remain within acceptable limits when compared to the natural variability observed in repeat 4DCT scans. Our model being agnostic to the DIR method chosen, the model can be improved by retraining using better DVFs, if available, to further refine the motion representation and enhance its accuracy and precision. Nevertheless, we recognize the importance of assessing the robustness of our method to variations in the chosen registration algorithm, as well as developing active mechanism to combat label noise. This aspect will be thoroughly investigated in future studies to ensure that our approach remains resilient to changes in registration methods.

Regarding respiration conditioning, we proposed to constrain the model with an amplitude-based scalar value extracted from patient-specific external 1D respiratory signal traces. To this end, we added a modified AdaIN layer to the bottleneck of the encoder based on this value. Doing so, we assumed a one-to-one correspondence between the amplitude and the 4DCT phase image, which represents a simplification of the problem. The use of AdaIN allows to consider more elaborate (e.g. higher dimensional) conditioning if available. For instance, more complex respiratory patterns could be better captured through clinically available real-time entire external surface measurements by optical devices \cite{nazir_technical_2018,ranjbar_validation_2021}. Including such information in the conditioning may thus further improve the model performance, which we will investigate in future research.

\section{Conclusion}
In this work, we have proposed a method to generate patient-specific pseudo respiratory motion with the clinical objective of reducing the use of 4DCT imaging in RT treatment planning. The model generates arbitrary breathing phases from static images by conditioning motion synthesis to a scalar value extracted from an external surface-based estimation. We train our generative model explicitly on state-of-the-art lung DVFs while encouraging motion realism under an adversarial objective that jointly constrains the warped image and the DVF magnitude.  

The method was extensively validated using two 4DCT lung tumor datasets showing different characteristics through a variety of images, organs at risk, and tumor level quality metrics.  
Results show for the first time that errors due to motion synthesis are not higher than differences between real repeat 4DCT acquisitions of the same patient, which is an important result towards 4DCT-free treatment planning. Future methodological developments will focus on more complex respiratory conditioning while further validation will evaluate the dosimetric impact of the proposed approach. 

\section*{References}
\newcommand{\newblock}{}

\bibliographystyle{plain}
\bibliography{Papier2023}
\end{document}